\documentclass[10pt,twocolumn,letterpaper]{article}
\usepackage{makecell}
\usepackage[usestackEOL]{stackengine}
\usepackage[pagenumbers]{cvpr} 

\definecolor{cvprblue}{rgb}{0.21,0.49,0.74}
\definecolor{forestgreen}{rgb}{0.13,0.54,0.13}
\usepackage[pagebackref,breaklinks,colorlinks,allcolors=cvprblue]{hyperref}

\def\paperID{*****} 
\def\confName{CVPR}
\def\confYear{2025}

\title{ReCapture: Generative Video Camera Controls for \\ User-Provided Videos using Masked Video Fine-Tuning}

\author{
  David Junhao Zhang$^{1,2}$ ~~~~~~~ Roni Paiss$^{1}$ ~~~~~~~ Shiran Zada$^{1}$ ~~~~~~~ Nikhil Karnad$^{1}$ ~~~~~~~ David E. Jacobs$^{1}$ \\ ~~ Yael Pritch$^{1}$ ~~~~~~~  Inbar Mosseri$^{1}$ ~~~~~~~  Mike Zheng Shou$^{2}$ ~~~~~~~ Neal Wadhwa$^{1}$ ~~~~~~~ Nataniel Ruiz$^{1}$\\
  $^{1}$Google\;\;\;\;   $^{2}$National University of Singapore \\
\\
\textcolor{blue}
{\href{https://generative-video-camera-controls.github.io/}{generative-video-camera-controls.github.io}}
}

\begin{document}

\twocolumn[{
\renewcommand\twocolumn[1][]{#1}
\maketitle
\begin{center}
    \centering
    \vspace*{-.5cm}
    \includegraphics[width=\linewidth]{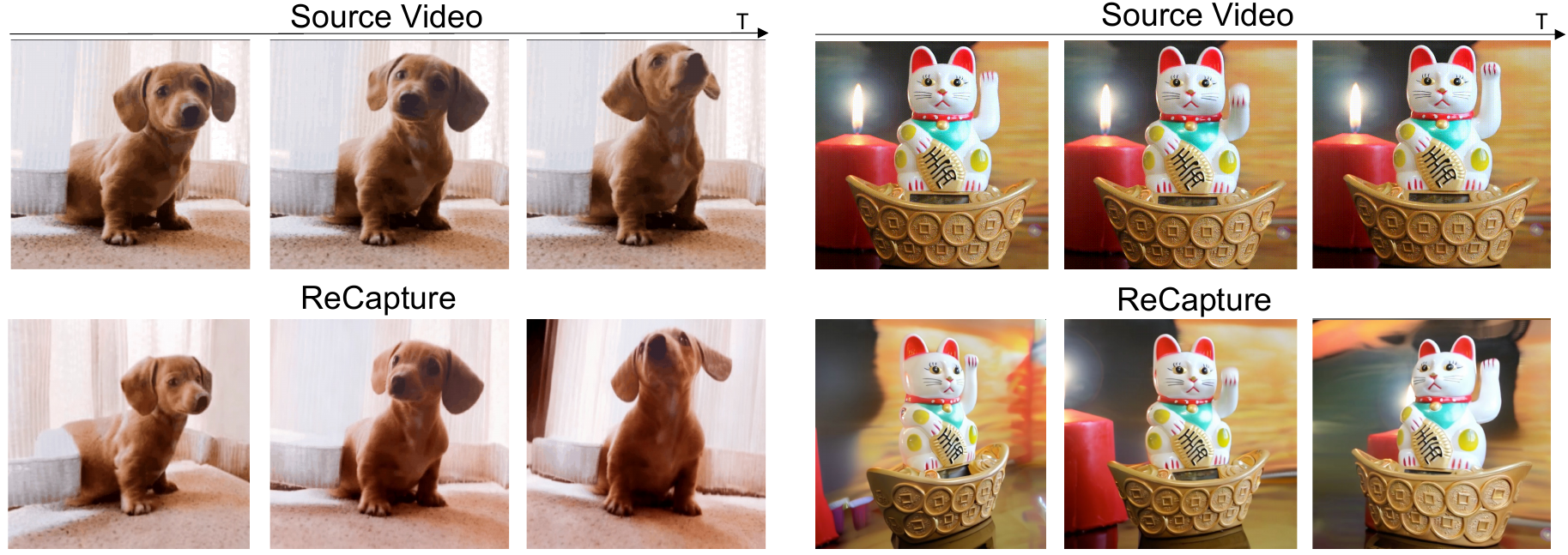}
    \captionof{figure}{
    Given a user-provided source video, using \textit{ReCapture}, we are able to generate a new version of the video with a new customized camera trajectory. Notice that the motion of the subject and scene in the video is preserved, and the scene is observed from angles that are not present in the source video.}
\label{fig:teaser}
\vspace*{-.1cm}
\end{center}
}]

\begin{abstract}
    Recently, breakthroughs in video modeling have allowed for controllable camera trajectories in generated videos. However, these methods cannot be directly applied to user-provided videos that are not generated by a video model. In this paper, we present \textit{ReCapture}, a method for generating new videos with novel camera trajectories from a single user-provided video.  Our method allows us to re-generate the reference video, with all its existing scene motion, from vastly different angles and with cinematic camera motion. Notably, using our method we can also plausibly hallucinate parts of the scene that were not observable in the reference video. Our method works by (1) generating a noisy anchor video with a new camera trajectory using multiview diffusion models or depth-based point cloud rendering and then (2) regenerating the anchor video into a clean and temporally consistent reangled video using our proposed masked video fine-tuning technique.
\end{abstract}    
\section{Introduction}
\label{sec:intro}

Recently, diffusion models have enabled significant advances in video generation and editing~\cite{ho2022imagen, singer2022make, he2022lvdm,zhang2023i2vgen,zhang2023show1,latentvideodiffusion,tuneavideo,blattmann2023align,magicvideo,qi2023fatezero,zhao2023motiondirector,menapace2024snap,yang2024cogvideox,gupta2023photorealistic}, revolutionizing workflows in digital content creation. Camera control plays a vital role in practical applications of video generation and editing, allowing for greater customization and stronger user experience. Recent efforts have introduced camera control capabilities to video diffusion models~\cite{he2024cameractrl,bahmani2024vd3d,motionctrl,hou2024training}, yet, in this case the videos are entirely generated by the video model from a text-prompt and are neither captured in the real world, nor provided by a user. Effectively generating new videos with user-specified camera motion from an existing user-provided video that contains complex scene motion is still an open and challenging problem.

The task is inherently ill-posed due to the limited amount of information in the reference video: one cannot know exactly how the scene looks like from all angles if there is not full knowledge of the scene's 4D content. However, this does not preclude an approximate solution that is plausible and appreciated by users. Previous studies have shown promising results by generally assuming the availability of synchronized multi-viewpoint videos~\cite{pumarola2020d} and constructing 4D neural representations. Later works~\cite{som2024,liu2023robust,lee2023casual-fvs,Wu_2024_CVPR,stearns2024dgmarbles} enable 4D reconstruction using a single monocular video, but require accurate camera pose and depth estimation, and cannot capture content outside the original field of view. In this paper, we reformulate this problem as a video-to-video translation task. Camera Dolly~\cite{van2024generative} also develops a video-to-video pipeline, but requires 4D video data with different camera poses obtained via simulation, which limits it to in-domain scenes like driving or cubic objects. Given the challenge of obtaining paired videos in the wild with varying camera movements, it is hard to solve this problem with a video-to-video pipeline in an end-to-end manner and we separate it into two steps instead. Our approach leverages the prior knowledge of diffusion generative models, in both image and video domains, to effectively reangle the video as if filmed from the requested camera trajectory. 

For the first stage of our method, we want to generate an incomplete anchor video conditioned on the user-provided camera trajectory and reference video. We initially obtain a partial frame-by-frame depth estimation of the target video. We project each frame into 3D space using a depth estimator to obtain a sequence of point clouds. Then, we simulate the user-specified camera movement, which can include zoom, pan, and tilt, and render the point cloud sequence according to the new camera trajectory. This estimation is only partial since, as illustrated in Figure~\ref{fig:intro_point}, these camera movements can introduce black areas outside the original video boundaries, cause some blurring due to the nature of point cloud projection and have poor temporal consistency since they are generated frame-by-frame. Another way to obtain the noisy anchor video, that uses recent advances in 3D reconstruction, is to use a multiview diffusion model~\cite{gao2024cat3d} conditioned on camera pose and individual video frames. This method also results in an anchor video that has poor temporal consistency, along with blurring, artifacts and black areas outside the scene.

Using this anchor video our method is able to generate a clean output with the desired camera trajectory. To achieve this we propose the novel technique of \textit{masked video fine-tuning}. This technique consists of training a context-aware spatial LoRA and temporal motion LoRA on the \textit{known} pixels from the generated anchor video, as well as from additional reference frame data. Specifically, the spatial LoRA is incorporated into the spatial layers of the video diffusion model and finetuned on augmented frames extracted from the source video. This enables the model to learn the subject's appearance and the background context of the source video. The temporal LoRA enables the model to learn the scene motion with respect to the new camera trajectory, and is inserted into the temporal layers of the video diffusion model and finetuned using a masked loss on the anchor video. Unknown regions are masked, which excludes them from the loss computation, enabling the model to focus on meaningful and known regions and motion while ignoring the unknown areas.

During inference, equipped with both video specific spatial and temporal LoRAs, the diffusion model can automatically fill the unknown regions of the anchor video with plausible content, leveraging the video diffusion model’s prior and the context provided by the spatial LoRA. It also significantly improves temporal consistency and removes anchor video jittering. This results in a coherent and meaningful video output, preserving the motion and layout of the original anchor video as learned through the temporal LoRA training. Finally, as a refining step, we can remove the temporal LoRA and retain only the context-aware spatial LoRA to apply SDEdit~\cite{meng2021sdedit} to the generated video, thereby further reducing blurring and improving temporal consistency.

In the end, we generate a video with new camera trajectories while preserving the original complex scene motion and the full content of the source video. Notably, this is accomplished without the need for paired video data. Ultimately, our method outperforms the generative approach  Generative Camera Dolly~\cite{van2024generative},  which requires paired videos as training data, and other  4D reconstruction methods~\cite{wu20244d,li2023dynibar}   on the Kubric dataset~\cite{van2024generative}. Furthermore, each component of our proposed method is validated through  ablation studies on VBench~\cite{huang2023vbench}.

\section{Related Work}

\begin{figure*}
    \centering
    \includegraphics[width=\linewidth]{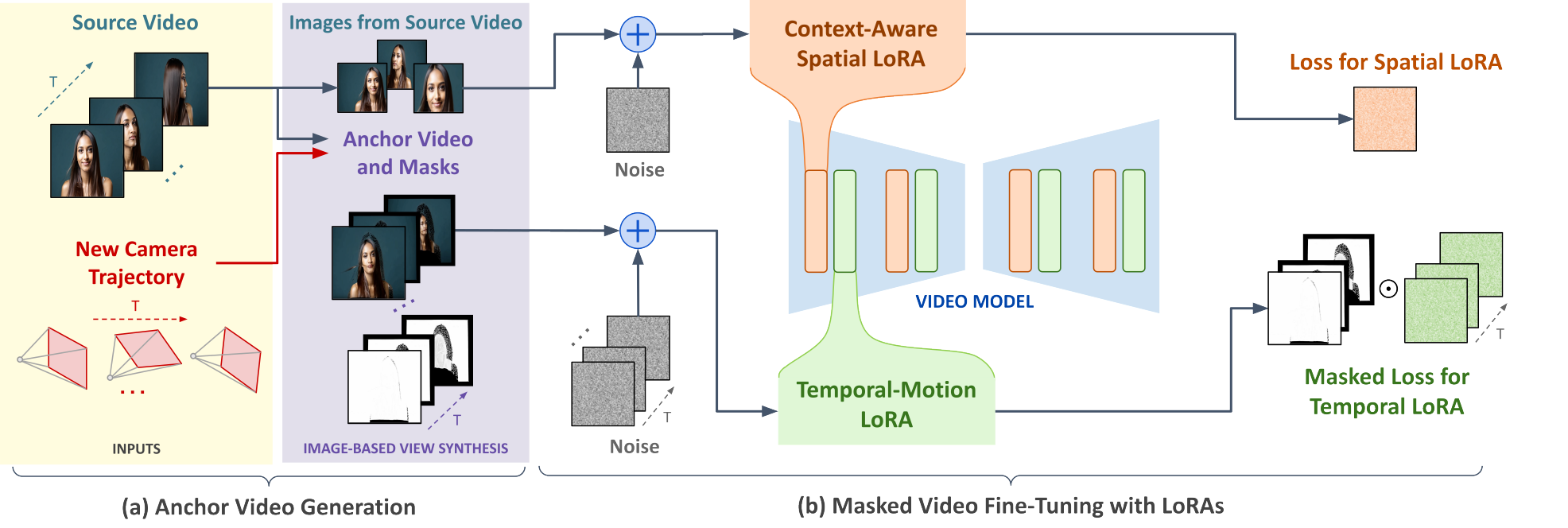}
    \caption{\textbf{ReCapture} consists, at setup time, of (a) Anchor video generation (b) Masked video fine-tuning using spatial and temporal LoRAs. To generate the clean output video with the new camera trajectory we simply perform inference of the video model.}
    \label{fig:intro}
\end{figure*}

\textbf{Video Diffusion Model.} Recent video generation methods \cite{karras2023dreampose, ruan2023mm, zhang2023i2vgen, he2022latent, chen2023videocrafter1, hong2022cogvideo,bar2024lumiere,chen2024videocrafter2,xing2025dynamicrafter,girdhar2023emu}, predominantly utilize diffusion models \cite{ho2020denoising, song2020denoising, peebles2023scalable} due to their state-of-the-art performance and robust open-source communities. 
Some approaches use a 3D-UNet architecture, inflating an image diffusion model with trainable temporal layers \cite{hong2022cogvideo,blattmann2023align,animatediff,bar2024lumiere,Blattmann2023StableVD}. Other works explored the transformers architecure, with both discrete \cite{kondratyuk2023videopoet} and continous tokens \cite{ma2024latte,gupta2023photorealistic,videoworldsimulators2024}.
In this paper, we use the open-source Stable Video Diffusion (SVD)~\cite{Blattmann2023StableVD} as our video diffusion model. 

\noindent\textbf{Personalization of Video Diffusion Models.}
At this stage the problem of personalization of image generative models has been well explored in the last several years, with work on subject-driven generation~\cite{ruiz2023dreambooth,Gal2022AnII,ruiz2024hyperdreambooth,chen2024subject,ye2023ip}, style-driven generation~\cite{sohn2023styledrop,wang2024instantstyle,hertz2024style,rout2024semantic}, style+subject-driven generation~\cite{shah2025ziplora,ruiz2024magic} and image-level personalization for inpainting~\cite{tang2024realfill}. The research direction of personalization of video models is more sparse, albeit with important recent work such as Dreamix~\cite{molad2023dreamix} which proposed to finetune video models on a given video, Still-Moving~\cite{chefer2024still} which mitigates the need for customized video data by elevating a customized image models to the video domain using spatial and temporal adapters and Movie Gen~\cite{polyak2024movie} which proposes directly training conditioning pathways for video models. Our method targets a wholly different application than this body of work, although these methods are important and related.

\noindent\textbf{Video Generation with Camera Control.}
Recent works studied adding explicit camera control to video generation models, in order to produce videos that both match the input text and align with a specified camera trajectory. 
Most methods~\cite{motionctrl,he2024cameractrl,bahmani2024vd3d,xu2024camco,kuang2024cvd} introduce additional modules that accept camera parameters, and train in a supervised manner. Other approaches, such as Viewcrafter~\cite{yu2024viewcrafter} and Training-Free \cite{hou2024training}, leverage rendering priors as input to achieve camera control.

Unlike these prior works, which incorporate camera control into the video generation process, our method alters the camera trajectory of user provided videos, while preserving the original content and dynamics.

\noindent\textbf{Novel View Synthesis of Dynamic Scene.} 
Novel View Synthesis has been a long standing task, where given a few sparse view of a scene the model predicts a novel view from an unobserved camera position ~\cite{liu2023zero,poole2022dreamfusion,lin2023magic3d,wang2023score,wang2023prolificdreamer,chen2023fantasia3d,haque2023instruct,po2023compositional,zhang2023scenewiz3d,hollein2023text2room,wu2023reconfusion,voleti2024sv3d}. 
However, extending such methods to videos is highly non-trivial since they rely on multiview training data. While obtaining different views of a scene difficult, it is significantly more feasible than obtaining synchronized video pairs with varying camera trajectories.

Recently, methods that generate novel view of dynamic scenes (i.e. videos) have been introduced. Initially, such methods relied on multiple synchronized input videos~\cite{bansal20204d, bemana2020x, li2022neural}, which limited its practical application. 
With the invention of NeRF~\cite{mildenhall2021nerf}, newer methods built a 4D volumetric representation of the scene and rendered different views using using time-evolving NeRFs~\cite{li2021neural, xian2021space, du2021neural, li2021neural, pumarola2021d, park2021nerfies}.
These methods requires the input video to include various camera views and struggles to generalize outside of the field of view present in the original video.

Recent approaches support more natural monocular videos. DynIBaR~\cite{li2023dynibar} employs a volumetric image-based rendering framework that aggregates features from nearby views from a monocular video in a camera-aware manner. DpDy~\cite{wang2024diffusion} leverages an image-based diffusion model to create a hybrid 4D representation, combining static and dynamic NeRF components. More recently, some approaches have employed dynamic 3D Gaussian splatting \cite{wu20234d, duan20244d} to reconstruct 4D scenes and achieve real-time rendering. 

While, generating appealing results, these methods often rely on effective multi-view cues generated by camera motion, which requires camera teleportation or quasi-static scenes and may be lacking in monocular videos \cite{gao2022monocular}. In addition, they often struggle to extrapolate outside of the provided field of view. Recently, several text-to-4D and image-to-4D papers have emerged; however, their results are mostly confined to animations of individual objects or animals~\cite{singer2023text,bahmani20244d,ling2024align,zhao2023animate124,ren2024l4gm,ren2023dreamgaussian4d,zhang20244diffusion}. Other studies~\cite{yu20244real,watson2024controlling} address more complex scenes, but they require 4D training data or sophisticated 4D reconstruction.

Instead of building an explicit 4D representation, our method harnesses the motion prior of video generation models, and reformulates the task as video-to-video translation (regenerating a given video from the requested camera trajectory). We show that our method performs effectively on real-world videos and generate meaningful content even when the camera moves beyond the original field of view.
\section{Method}

Our method works in two stages, first by generating an incomplete and noisy anchor video with respect to the new camera trajectory and second by regenerating this anchor video into a clean and temporally consistent reangled video using our proposed \textit{masked video fine-tuning} technique. 

In more detail, the first stage consists of image-based view synthesis, in which we independently transform each input video frame to produce noisy anchor frames with the new camera pose, along with their validity masks. These frames are typically incomplete; they have artifacts such as missing information from revealed occlusions, and have structural deformations and temporal inconsistencies such as flickering.

Our overall method is agnostic to the specific technique used to generate the anchor frames in the first stage, and in this work we explore two different techniques: point-cloud sequence rendering, and multi-view per-frame image diffusion. Our full method can correct errors in the anchor video, solve temporal inconsistency and complete missing information.

In the second stage we apply a novel masked video fine-tuning strategy. We train a temporal motion LoRA on the anchor video with a masked loss (that masks the corrupted parts of the anchor video). This directs the model to generate a video that follows the camera motion and dynamics of the anchor video while completing missing parts according to the model prior. In addition, we train a context-aware spatial LoRA on augmented images generated from frames of the \textit{source video}. This is done by disabling the temporal layers of the video model allowing us to train on images. This step allows the model to fix structural artifacts in the anchor video. After fine-tuning, the video diffusion model is able to regenerate the anchor video such that the missing areas are  seamlessly filled with temporally and spatially consistent video content. Furthermore, it preserves the coherence, content and dynamics of the original video and eliminates artifacts from the first stage, all while adapting to the new camera motion. The final output is simply the source video with the new camera trajectory, which often includes views of the scene that are unseen in the source video.

\begin{figure}
    \centering
    \includegraphics[width=\linewidth]{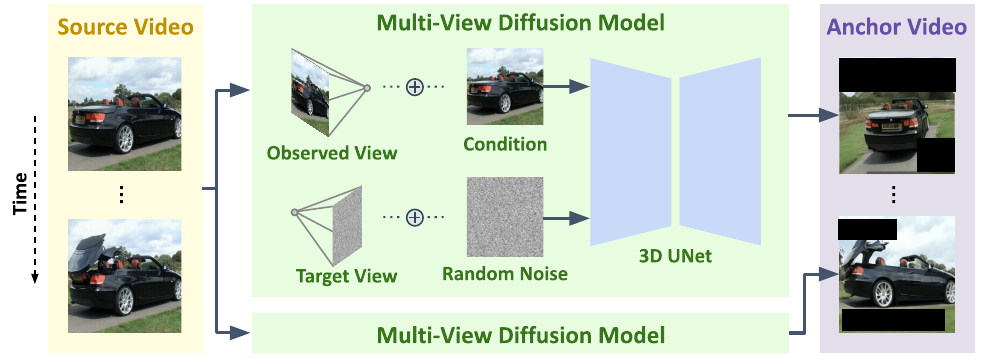}
    \caption{\textbf{Anchor video generation} using image-level multiview-diffusion models to generate new views frame-by-frame.}
    \label{fig:intro_multiview}
\end{figure}

\begin{figure}
    \centering
    \includegraphics[width=\linewidth]{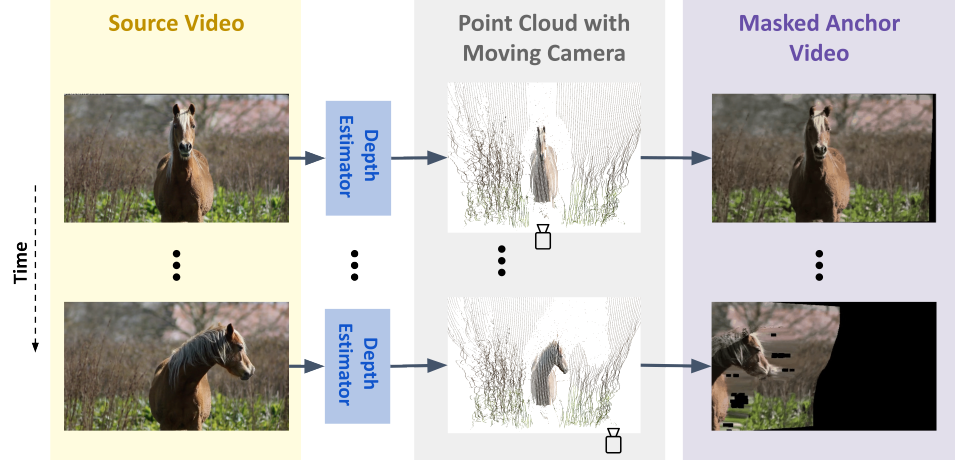}
    \caption{\textbf{Anchor video generation} using depth estimation to turn each frame into a point cloud and then generating new views by controlling the camera pose. }
    \label{fig:intro_point}
\end{figure}

\subsection{Preliminaries}
\textbf{3D U-Net.}
Video diffusion models train a 3D U-Net (spatial-temporal for 3D) to denoise a sequence of Gaussian noise samples to generate videos, guided by text or image prompts~\cite{blattmann2023align,blattmann2023stable}. The 3D U-Net architecture comprises down-sampling, middle, and up-sampling blocks. Each block includes multiple spatial convolution layers, spatial transformers, cross-attention layers, and either temporal transformers or convolution layers.

\noindent\textbf{Low-Rank Adaptation (LoRA).} Low-rank adaptation was introduced to fine-tune large pre-trained language models~\cite{hu2021lora} and has since been applied to text-to-image and text-to-video tasks for appearance customization~\cite{ruiz2023dreambooth,chefer2024still}. It updates the weight matrix $W$ using low-rank factorization as:

\begin{equation}
W = W_0 + \Delta W = W_0 + BA,
\end{equation}

where $W_0$ represents the original weights, and $B$ and $A$ are low-rank factors with much fewer dimensions. LoRA is computationally efficient and can be more easily regularized, which preserves more of the model’s prior knowledge compared to fine-tuning the entire network.
\subsection{Anchor Video with New Camera Motion}\label{sec:anchorvideo}

At this stage, we are given a reference video $\mathbf{V} = \left [ \mathbf{I}_{0}, \dots, \mathbf{I}_{N-1} \right ] \in \mathbb{R}^{N \times 3 \times H \times W}$, where $N$ represents the number of frames. The main objective is to transform these frames into a new sequence, denoted as $\mathbf{V}^{a}$, based on a different camera trajectory provided by the used. We refer to this video as an \textit{anchor} because it anchors the final output of our method and serves as a condition for the next stage. This first draft contains artifacts from out-of-scene regions and temporal inconsistencies, which will be corrected in the second stage.

We present two approaches to obtain an anchor video based on a new camera trajectory. The first method uses a point cloud rendering technique, suitable for typical camera movements such as panning, tilting, and zooming, which involve simple translations and small rotations. The second approach is designed for camera movements that involve larger rotations such as orbiting. In this approach, we utilize a multiview diffusion model~\cite{gao2024cat3d} to generate novel views of the scene.

\noindent\textbf{Point Cloud Sequence Rendering.}
We begin by lifting the pixels from the input image plane into a 3D point cloud representation. For each frame of the source video $\mathbf{I}_{i}$, $i \in \{ 0, ..., N-1 \}$, we independently estimate its depth map $\mathbf{D}_i$ using an off-the-shelf monocular depth estimator~\cite{bhat2023zoedepth}. By combining the image with its depth map, the point cloud $\mathcal{P}_i$ can be initialized as:

\begin{equation} \mathcal{P}_i = \phi([\mathbf{I}_i, \mathbf{D}_i], \mathbf{K}), \end{equation}

where $\phi$ denotes the mapping function from RGBD to a 3D point cloud in the camera coordinate system, and $\mathbf{K}$ represents the camera's intrinsics using the convention in~\cite{chung2023luciddreamer}.

Next, we take as input the camera motion as a pre-defined trajectory of extrinsic matrices $\left \{\mathbf{P}_1, ... , \mathbf{P}_{N-1} \right \}$, where each includes a rotation matrix and a translation matrix representing the camera's pose (position and orientation), which are used to rotate and translate the point cloud in the camera's coordinates.

We then project the point cloud of each frame back onto the anchored camera plane using the function $\psi$ to obtain a rendered image with perspective change: $\mathbf{I}^{a}_{i} = \psi (\mathcal{P}_i, \mathbf{K}, \mathbf{P}_i)$. By calculating the extrinsic matrices corresponding to the camera's movement, we can express a variety of camera motions including zoom, tilt, pan, pedestal, and truck, enabling flexible camera control to yield anchor videos:

 \begin{equation} 
 \resizebox{0.9\columnwidth}{!}{$
   \mathbf{V}^{a} = \left \{ \mathbf{I}^{a}_0, ... , \mathbf{I}^{a}_{N-1} \right \} = \left \{ \psi (\mathcal{P}_i, \mathbf{K}, \mathbf{P}_i) | i \in \left \{ 0, ..., N-1 \right \} \right \} .
   $}
\end{equation}

Simultaneously with the color frames, we obtain a binary mask for each frame. Valid pixels after projecting the point cloud, represented with a value of '1'. Regions missing due to camera movement as shown in Fig.~\ref{fig:intro_point}, which extend beyond the original video scene, are marked as '0'. We denote the corresponding sequence of binary masks as $\mathbf{M}^a \in \mathbb{R}^{N \times 1 \times H \times W}$.

\begin{figure*}[t]
    \centering
    \includegraphics[width=\linewidth]{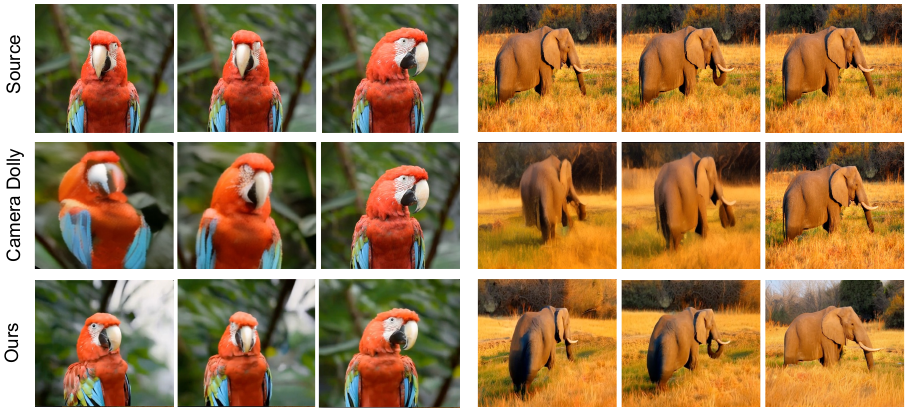}
    \caption{Comparisons with generative camera dolly~\cite{van2024generative} using an \textit{orbit} camera trajectory. }
    \label{fig:intro_comparsions}
\end{figure*}

\noindent\textbf{Multiview Image Diffusion for Each Frame.}  When a camera trajectory involves significant rotation and viewpoint changes, point cloud rendering usually fails~\cite{yu2024viewcrafter}. To address this, we employ a multiview diffusion model~\cite{gao2024cat3d}. This approach leverages the fact that multiview image datasets are generally easier to obtain compared to multiview video datasets. Specifically, as shown in Fig.~\ref{fig:intro_multiview}, for each frame $\mathbf{I}_i$ of the source video, which represents the condition view, along with its corresponding camera parameters $\mathbf{P}_{\text{cond}}$, the model learns to estimate the distribution of the target image $\mathbf{I}^{a}_i$
\begin{equation}
    p\left(\mathbf{I}_{i}^a \mid \mathbf{I}_{i}, \mathbf{P}_{cond}, \mathbf{P}_{i} \right)\,. 
\end{equation}
where $\mathbf{P}_i$ is the target camera parameters which are also provided as input.
The multiview diffusion model employs a 3D U-Net as its backbone, and for simplicity of notation, we omit the latent VAE encoder. The 3D U-Net is adapted from a 2D text-to-image U-Net by inflating the 2D self-attention mechanism into 3D, allowing it to operate across both 2D spatial dimensions and multiple view images. Due to the difficulty in obtaining the camera pose with the condition frame, we follow CAT3D and use a raymap with the same dimensions as the condition images, computed relative to the first image's camera pose. This makes the pose representation invariant to rigid transformations. Raymaps are then concatenated channel-wise to the corresponding condition image. 

By applying this multiview diffusion model to each frame, we can  generate the anchor video $\mathbf{V}^{a}$. However, processing each frame independently leads to significant temporal inconsistencies and artifacts (Fig.~\ref{fig:intro_point}).
While CAT3D can complete unseen regions, it hallucinates these region differently at each frame. To mask out this regions, we obtain the mask $\mathbf{M}^a$ with the same approach as point cloud rendering, which indicate the regions missing due to added camera movement.

\subsection{Masked Video Fine-tuning}

At this stage, our objective is to take an incomplete anchor video (with significant artifacts and inconsistencies) as input and produce a clean, high-quality output. To achieve this, we introduce the technique of \textit{masked fine-tuning} of a video diffusion model, incorporating a context-aware spatial LoRA and a temporal motion LoRA. By leveraging the strong prior of the video diffusion model, we can generate a high-quality video conditioned on the anchor video. Next we present our components:

\noindent\textbf{Temporal LoRAs  with Masked Video Fine-tuning.} The anchor video from the first stage may exhibit significant artifacts, such as revealed occlusions due to camera movement and temporal inconsistencies such as flickering. To address these issues, we propose a masked video fine-tuning strategy using temporal motion LoRAs. LoRAs are applied to the linear layers of the temporal transformer blocks in the video diffusion model. Since LoRA operates in a low-rank space and the spatial layers remain untouched, it focuses on learning fundamental motion patterns from the anchor video without over-fitting to the entire video. The strong temporal consistency prior from the video diffusion model helps minimize temporal inconsistencies. We introduce a masked diffusion loss, where the invalid regions in the anchor video are excluded from the loss calculation, ensuring the model only learns from meaningful pixels. During inference, the video diffusion model regenerates the video and automatically fills in the invalid regions while maintaining the original motion of the anchor video. 
The temporal loss for diffusion training is defined as:

\begin{equation}
    \mathcal{L}_{temp} = \mathbb{E}_{\epsilon, t}\left \lbrack  
    \mathbf{M}^a \cdot\lVert \epsilon - \epsilon_\theta(\mathbf{V}^{a}_{t}, t, y) \rVert \right\rbrack,
\end{equation}

where $\epsilon$ represents noise added to the anchor video $\mathbf{V}^a$, $\mathbf{V}^a_t$ denotes the noisy anchor video at time step $t$, $y$ refers to the text or image condition, and $\theta$ indicates the weights of the 3D U-Net along with the LoRA weights.

\begin{figure*}[p]
    \centering
    \includegraphics[width=0.9\linewidth]{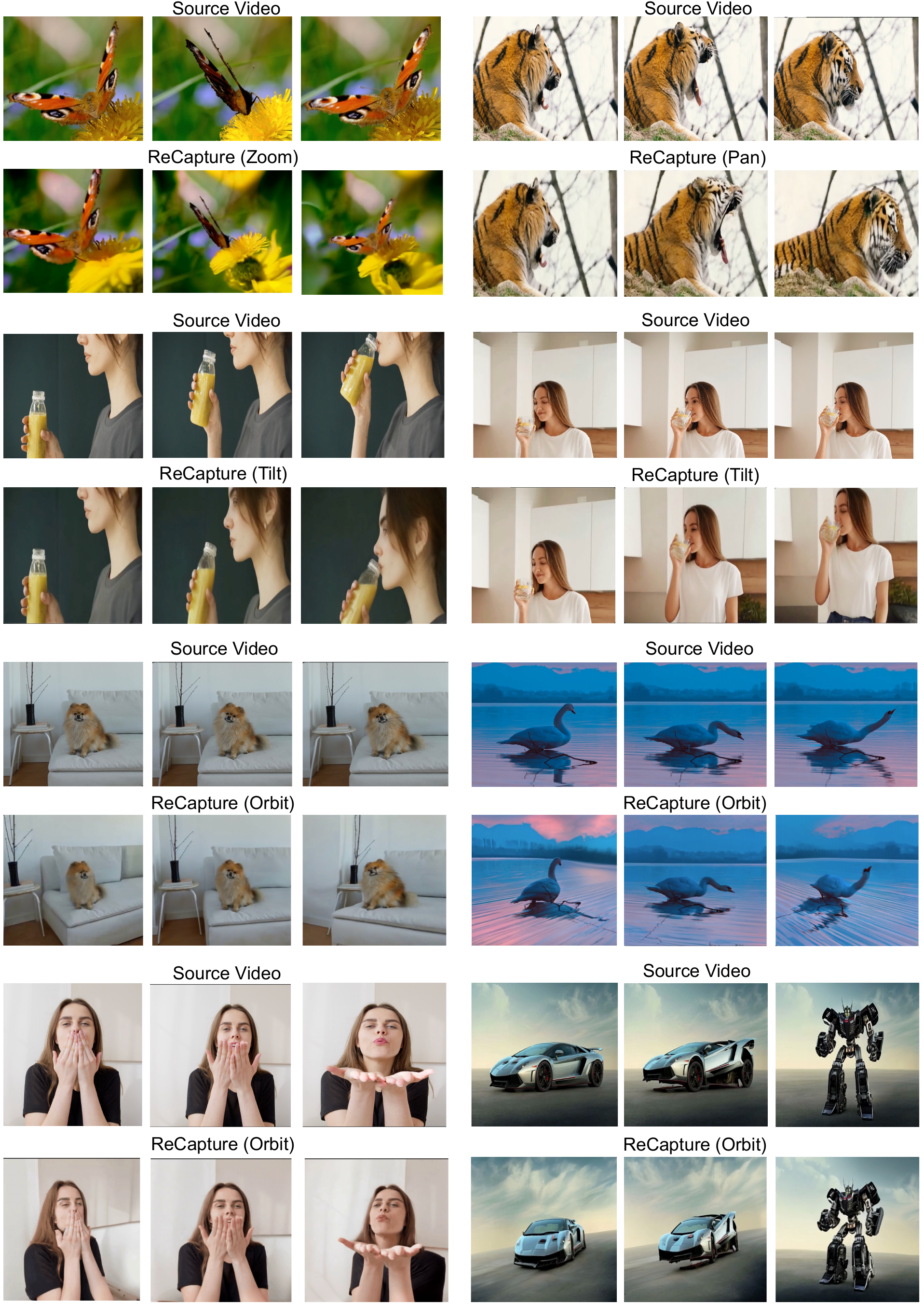}
    \caption{\textbf{Gallery} of generated videos with novel and unseen user-provided camera trajectories using \textit{ReCapture}. }
    \label{fig:gallery}
\end{figure*}

\noindent\textbf{Context-Aware Spatial LoRAs.}
Although the video diffusion model with masked fine-tuning automatically fills the invalid regions of the anchor video, the filling may not be consistent with the original context or appearance, and might appear pixelated, as shown in Fig.~\ref{fig:ABLATION2} Line 2. To address this issue, we propose enhancing the spatial attention layers of the video diffusion model by incorporating a spatial LoRA, which is fine-tuned on the frames of the source video. At each training step, a frame is randomly selected from the source video, and the temporal layers are bypassed. The spatial LoRA loss is defined as follows
\begin{equation}
    \mathcal{L}_{spatial} = \mathbb{E}_{\epsilon, t, i \sim\mathcal{U} \left\{ 0,... \mathit{N}-1\right\}}\left\lbrack 
    \lVert \epsilon - \epsilon_\theta((\mathbf{I}_{i,t}), t, y) \rVert \right\rbrack,
\end{equation}
where $\mathbf{I}{i,t}$ denotes the noisy frame $\mathbf{I}{i}$ of the source video at time step $t$. The spatial LoRA captures the original context from the source video, ensuring seamless integration of filled pixels with the original pixels.

Consequently, our final diffusion loss is the sum of $L_{temp}$ and $L_{spatial}$. To ensure compatibility between the spatial and temporal LoRAs, features from the corrupted video for training temporal LoRA are also passed through the spatial LoRA, without updating its parameters.

\begin{table*}[t]
\centering
 \scalebox{0.90}{
\begin{tabular}{c|c|c|c|c|c|c|c|c}
\hline
Models & \Centerstack{Subject\\Consistency} & \Centerstack{Background\\Consistency} & 
\Centerstack{Temporal\\Flickering} & 
\Centerstack{Motion\\Smoothness} & 
\Centerstack{Dynamic\\Degree} & 
\Centerstack{Aesthetic\\Quality} &  
\Centerstack{Imaging\\Quality}&
\Centerstack{Object\\Class}\\ \hline

Generative Camera Dolly~\cite{van2024generative} & 83.02\% & 80.42\% & 74.64\% & 82.33\% & \textbf{51.24\%} & 38.67\% & 58.62\% & 76.46\% \\ 

Ours& \textbf{88.53}\% & \textbf{92.02}\% & \textbf{91.12}\% & \textbf{98.24}\% & 49.03\% & \textbf{57.35}\% & 64.75\% & \textbf{82.07}\%  \\
\hline

\end{tabular}}
\caption{\textbf{Quantitative comparisons} with Generative Camera Dolly on VBench.}
\label{tab:semantic_evaluation}

\end{table*}

\begin{table}
\resizebox{\columnwidth}{!}{
  \centering
  \scalebox{0.95}{
  \begin{tabular}{@{}l|rrrrr@{}}
    \toprule
    Method & \thead{PSNR\\ (all)~$\uparrow$} & \thead{SSIM\\ (all)~$\uparrow$} & \thead{LPIPS\\ (all)~$\downarrow$} & \thead{PSNR\\ (occ.)~$\uparrow$} & \thead{SSIM\\ (occ.)~$\uparrow$} \\
    
    \midrule
    HexPlane~\cite{cao2023hexplane} & 15.38 & 0.428 & 0.568 & 14.71 & 0.428 \\
    4D-GS~\cite{wu20244d} & 14.92 & 0.388 & 0.584 & 14.55 & 0.392 \\
    DynIBaR~\cite{li2023dynibar} & 12.86 & 0.356 & 0.646 & 12.78 & 0.358 \\
    
    \midrule
    Vanilla SVD~\cite{blattmann2023stable} & 13.85 & 0.312 & 0.556 & 13.66 & 0.326 \\
    ZeroNVS~\cite{sargent2023zeronvs} & 15.68 & 0.396 & 0.508 & 14.18 & 0.368 \\

   Generative Camera Dolly~\cite{van2024generative} & 20.30 & 0.587& 0.408 & 18.60 & 0.527 \\ %
    
    \midrule
    Ours & \textbf{20.92} & \textbf{0.596} & 0.402 & \textbf{18.92} & \textbf{0.541} \\ %
  
  \bottomrule
  \end{tabular}
  }}
  \vspace{0.07in}
  \caption{
  \textbf{Comparison results on Kubric-4D.} We evaluate gradual dynamic view synthesis models following~\cite{van2024generative} to  use video with resolution $384\times 256$. Our method achieves superior performance compared to other reconstruction and generative methods.}
  \label{kubric_sota1}
\end{table}

\begin{table*}[t]
\centering
 \scalebox{0.9}{
\begin{tabular}{c|c|c|c|c|c|c|c|c}
\hline
Models & \Centerstack{Subject\\Consistency} & \Centerstack{Background\\Consistency} & 
\Centerstack{Temporal\\Flickering} & 
\Centerstack{Motion\\Smoothness} & 
\Centerstack{Dynamic\\Degree} & 
\Centerstack{Aesthetic\\Quality} &  
\Centerstack{Imaging\\Quality}&
\Centerstack{Object\\Class}\\ \hline
Anchor Video    & 82.41\% & 77.45\% & 64.50\% & 74.27\% & \textbf{49.72}\% & 34.94\% & 55.90\% & 79.82\% \\ 
+ Temporal LoRAs w/ Masks ) & 85.24\% & 90.88\% & 89.60\% & 97.32\% & 49.64\% & 40.41\% & 62.34\% & 80.02\% \\ 
++ Spatial LoRAs) & 86.02\% & 91.24\% & 90.02\% & 97.32\%  & 49.64\% & 49.18\% & 63.03\% & 80.02\% \\ 
+++ SD-Edit& \textbf{88.53}\% & \textbf{92.02}\% & \textbf{91.12}\% & \textbf{98.24}\% & 49.03\% & \textbf{57.35}\% & \textbf{64.75}\% & \textbf{82.07}\%  \\
\hline
\end{tabular}}
\caption{\textbf{Ablation studies} for each component of mask video diffusion finetuning: '+ Temporal LoRAs' applies temporal LoRAs solely for masked video finetuning. '++ Spatial LoRAs' introduces additional context-aware LoRAs, using both spatial and temporal LoRAs for finetuning. '+++ SD-Edit' involves applying SD-editing after completing training with both LoRAs for eliminating blurriness.}
\label{ablation}

\end{table*}

\begin{figure*}[t]
    \centering
    \includegraphics[width=\linewidth]{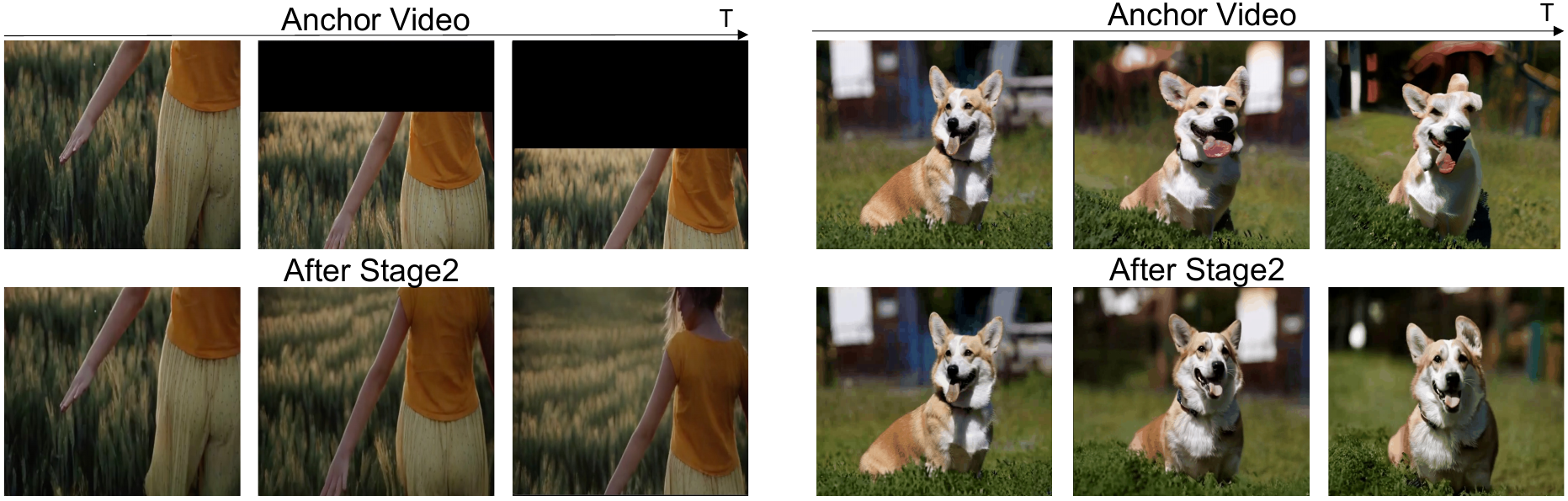}
    \caption{Visualization of the effectiveness of masked video fine-tuning (Stage 2) for generating spatially and temporally coherent outputs from noisy anchor videos. }
    \label{fig:ABLATION1}
\end{figure*}

\noindent\textbf{Eliminating blurriness.} After finishing training both LoRAs, we can directly use the video diffusion model to generate the desired video with new camera motion while maintaining high visual quality. Finally, as a post-processing stage to further eliminate blurriness, which cannot be fully addressed by masked video fine-tuning as shown in Fig.~\ref{fig:ABLATION2} Line 3, we use the video diffusion model with spatial LoRA while omitting the temporal LoRA to perform SD-editing~\cite{meng2021sdedit} on the output video. Typically, SD-editing introduces randomness that alters the appearance of the subject, but since our spatial LoRA has been fine-tuned on the source video, it preserves the original appearance while simultaneously removing the blur.

Our method successfully adds dynamic camera motion to the existing videos without the need for large-scale training on 4D multi-view video data, and it generalizes to a wide array of videos and camera trajectories. Next, we present our experiments, datasets, evaluations, and comparisons.

\section{Experiments}
We conduct a comprehensive qualitative and quantitative evaluation to assess the performance of our method, comparing it against prominent baselines for the task of novel view synthesis in existing videos. Our quantitative analysis in~\cref{subsecion:quantitative} encompasses two complementary subsets of automatic metrics: low-level statistics, including PSNR and SSIM, and high-level semantic measures, such as subject consistency. Additionally, we incorporate a user study for non-automatic evaluation. An ablation study in~\cref{subsecion:ablation_study} further demonstrates the importance of each component in our approach. In~\cref{subsecion:qualitative}, we present qualitative results that visually illustrate our method’s superiority, and implementation details are provided in~\cref{subsec:implementation_details}.

\subsection{Quantitative Evaluation}\label{subsecion:quantitative}

\textbf{Low-level evaluation metrics.} We utilize the Kubric-4D dataset, consisting of 3,000 scenes, including an evaluation subset of 100 scenes. These scenes, generated with the Kubric simulator, showcase complex multi-object interactions and dynamic movement patterns. Each scene includes synchronized video from 16 fixed camera viewpoints at a resolution of 576 x 384 across 60 frames at 24 FPS. Additionally, each scene contains 7 to 22 objects of varying sizes, with roughly one-third initially positioned mid-air to introduce intricate dynamics. This arrangement leads to frequent and complex occlusions, posing a substantial challenge for accurate novel view synthesis. For consistency, we adhere to the evaluation protocol outlined in \cite{van2024generative}, downsampling the video to 14 frames at a resolution of $384 \times 256$. 

As shown in Table~\ref{kubric_sota1}, our method outperforms existing 4D reconstruction methods, underscoring the effectiveness of framing camera control as a video-to-video translation task using generative models. Notably, our approach achieves superior results even without 4D training data, surpassing Generative Camera Dolly, which depends on extensive 4D data for training.

\noindent\textbf{High-level semantic evaluation metrics.}
As noted in~\cite{liu2023hosnerf}, low-level evaluation metrics often fall short in accurately capturing the true quality of a video. For example, while PSNR values may be similar, the visual quality of Dolly (as shown in \cref{fig:intro_comparsions}) is significantly blurrier and inferior to ours. To provide a more comprehensive and fair assessment of video quality, we conduct evaluations following the VBench dataset~\cite{huang2023vbench}. The benchmark comprises 35 videos and evaluates seven critical dimensions of video generation, including factors like subject identity consistency, motion smoothness, and temporal flickering. These fine-grained evaluation metrics employ feature extractors such as DINO~\cite{caron2021emerging} for assessing consistency and MUSIQ~\cite{ke2021musiq} for measuring image quality. This approach allows for a higher-level evaluation, offering deeper insights into each model’s distinct strengths and areas for improvement. 
We supply the specific prompt associated with each video to facilitate Object Class evaluation

As shown in Table~\ref{tab:semantic_evaluation}, our method outperforms in most evaluation dimensions by a large margin. This approach allows for a higher-level evaluation, offering deeper insights into each model’s distinct strengths and areas for improvement.

\subsection{Qualitative Comparisons}\label{subsecion:qualitative}
For qualitative results, we compare our method with Generative Camera Dolly~\cite{van2024generative}, as shown in Figure 4. The comparison demonstrates that our method more accurately follows the camera trajectory, generates fewer artifacts, and exhibits less blurriness than Generative Camera Dolly. Notably, even when the camera position differs significantly from the original video, our method faithfully preserves the subject's motion and appearance, whereas Dolly suffers from significant artifacts under these conditions.

\subsection{Ablation Studies.}\label{subsecion:ablation_study}
\begin{figure}[t]
    \centering
    \includegraphics[width=\linewidth]{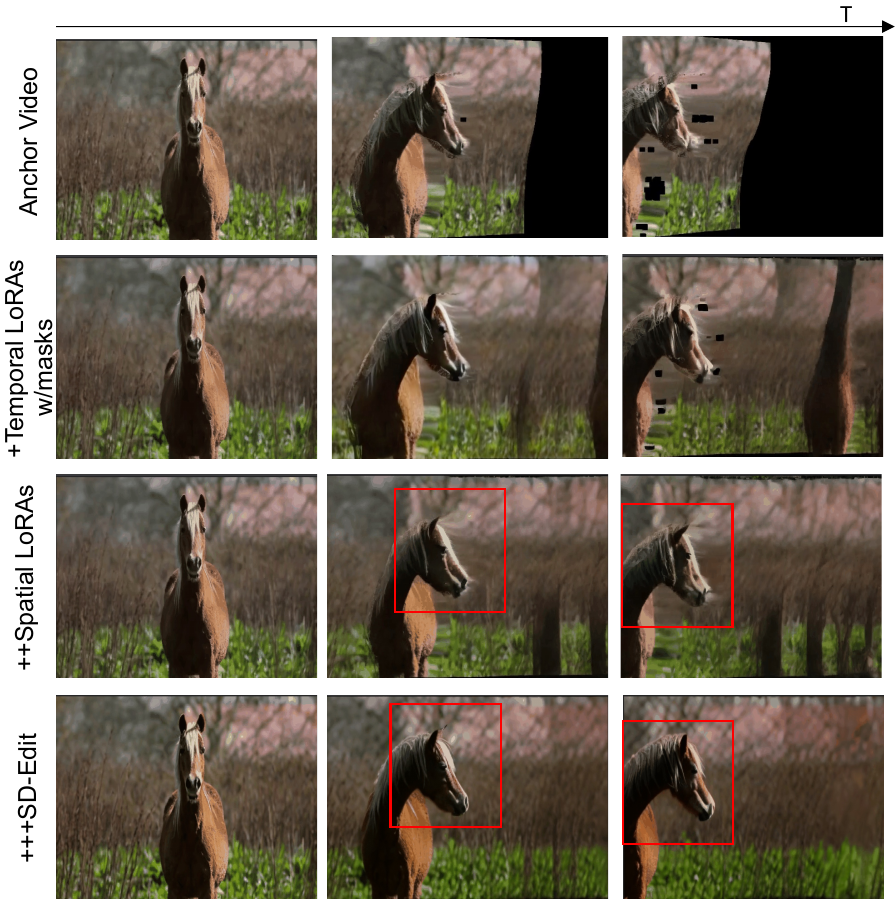}
    \caption{Detailed ablation of all components of our method. }
    \label{fig:ABLATION2}
\end{figure}

We evaluate the impact of each component of our method in Table~\ref{ablation}. As demonstrated, fine-tuning the temporal LoRA with masking significantly improves temporal consistency and visual quality. This effect is also visible in Fig.~\ref{fig:ABLATION2}, where generating the video with temporal LoRA helps to fill regions not visible in the anchor video, enhancing both visual quality and temporal coherence. Additionally, incorporating the context-aware spatial LoRA further boosts visual quality and subject consistency by leveraging prior knowledge of the original video. Finally, integrating SDEdit further enhances overall quality by reducing artifacts. These results affirm the effectiveness of each component in our method.

\subsection{Implementation Details.}\label{subsec:implementation_details}

We use the the CAT3D~\cite{gao2024cat3d} multi-view model without any further adjustments. We employ SVD~\cite{svd} as our video diffusion model in all our experiments, since it is I2V model, we use $\mathbf{V}^a_0$ as the image prompt. For the LoRA finetuining we use rank of 16 for both spatial and temporal LoRAs. The spatial LoRA is added to the self-attention layers, while the temporal LoRA is integrated into the temporal attention layers, with the learning rate set to $5e^{-4}$.
The total number of fine-tuning steps is set 400 and requires 5~min on single 80GB A100 GPU.

\section{Conclusion}

In this work we present \textit{ReCapture}, our proposed method to generate new videos with novel camera trajectories from existing user-provided videos. Compared to previous work, and thanks to the strong prior of video models, ReCapture has a surprisingly strong ability to generalize to vastly different videos and scenes, and in many cases faithfully preserves complex scene and subject motion, as well as the details of the scene.


{
    \small
    \bibliographystyle{ieeenat_fullname}
    \bibliography{main}
}

\end{document}